\documentclass[preprint,review,12pt]{elsarticle}
\usepackage{amssymb, amsmath, amsthm, bm}
\usepackage[noend]{algpseudocode}
\usepackage{algorithmicx,algorithm}
\usepackage{graphicx}
\usepackage{caption}
\usepackage[colorlinks,linkcolor=blue]{hyperref}
\usepackage{epstopdf}
\usepackage[colorlinks,linkcolor=blue]{hyperref}
\usepackage{bbding}
\usepackage{booktabs}
\usepackage{tikz,xcolor}% Make Orcid icon
\usepackage[caption=false]{subfig}
\usepackage{multirow}
\usepackage{makecell}
\usepackage{bm} % optional
\usepackage{algorithmicx,algorithm}
\usepackage{diagbox}
\usepackage{listings}

\newtheorem{rmrk}{Remark}
\numberwithin{equation}{section}

\journal{}

\begin{document}
\begin{frontmatter}
\title{Missingness Augmentation: A General Approach for Improving Generative Imputation Models }
\author[label1] {Yufeng~Wang\fnref{cor1}}
\author[label1] {Dan~Li\fnref{cor2}}
\author[label1] {Cong~Xu\fnref{cor3}}
\author[label1] {Min~Yang\corref{cor4}}
\fntext[cor1] {Email: ytuyufengwang@163.com }
\fntext[cor2] {Email: danliai@hotmail.com}
\fntext[cor3] {Email: congxueric@gmail.com}
\cortext[cor4] {Corresponding author: yang@ytu.edu.cn}

\address[label1]{School of Mathematics and Information Sciences, Yantai University, Yantai, China}

\begin{abstract}
Missing data imputation is a fundamental problem in data analysis,
and many studies have been conducted to improve its performance by exploring model structures and learning procedures.
However, data augmentation, as a simple yet effective method, has not received enough attention in this area.
In this paper, we propose a novel data augmentation method called Missingness Augmentation (MisA) for generative imputation models.
Our approach dynamically produces incomplete samples at each epoch by utilizing the generator's output,
constraining the augmented samples using a simple reconstruction loss, and combining this loss with the original loss to form the final optimization objective.
As a general augmentation technique,
MisA can be easily integrated into generative imputation frameworks, providing a simple yet effective way to enhance their performance.
Experimental results demonstrate that MisA  significantly improves the performance of many recently proposed generative imputation models on a variety of tabular and image datasets.
The code is available at \url{https://github.com/WYu-Feng/Missingness-Augmentation}.
\end{abstract}

	
\end{frontmatter}

\section{Introduction}
Machine learning requires substantial amounts of elaborate data to achieve compelling performance,
but incomplete data are more pervasive than fully-observed data in practice.
An effective way to tackle missing data is to perform missingness imputation \cite{Fais2022,Fig2023,Little2019}.
Current imputation methods can be categorized into two types:
discriminative methods and generative ones.
Benchmark discriminative models include MICE \cite{Buuren2011}, MissForest \cite{Stekhoven2012},
Matrix Completion \cite{Mazumder2010}, and so on.
However, due to the reliance on the assumption that there exists certain linear correlation between the missing and the observed parts,
the performance of discriminative models is unsatisfactory for complex data.
Compared to discriminative models,
generative models, especially deep generative models, can better capture the underlying nonlinear latent structure among the incomplete data.
In recent years,
there has been a surge of interest in developing various deep generative models to better estimating missing or corrupted data;
see e.g. \cite{Gondara2018, Li2019, Lin2022,Nazabal2020, Richardson2020, Sam2022,Yoon2018}.
In spite of the promising advances brought by these deep imputation models,
the effectiveness of the models depends heavily on high-quality and diverse trainable data.
Limited data availability can cause a significant decline in the imputation accuracy of the model, further resulting in poor performance in downstream tasks.

Data augmentation,
either by slightly modifying existing data instances \cite{Cubuk2019, Cubuk2020, Zhong2020},
or by generating synthetic instances from existing ones \cite{Devires2017, Kuo2020, Li2021,Yun2019},
is considered as one of the effective ways to address the lack of trainable data and improve the generalization ability of models.
The benefits of data augmentation have been widely acknowledged in many machine learning tasks
in computer vision \cite{Li2023,Minaee2021} and natural language processing \cite{Feng2021}.
However, the augmentation methods that can be used for missing data imputation are still lacking,
primarily due to two reasons.
Firstly, the missingness of data makes it  difficult to construct samples that are consistent with the target distribution domain.
The augmented samples are highly likely to deviate from the actual data distribution, particularly when the missing rate is high.
Secondly, the current data augmentation techniques often implicitly rely on statistical correlations between attributes, such as pixels of the image or words in a sentence.
Nonetheless, for large amounts of practical tabular data,
there is no such spatial or sequential correlation between attributes,
rendering these techniques unsuitable for missing tabular data.

In this paper, we propose a simple, efficient but effective data augmentation approach
that can be easily plugged into many deep generative missing data imputation models.
Intuitively, we posit that a good imputation model should not only use observations to infer missing values
but also produce imputed values that accurately reflect the raw observations.
Motivated by this intuition, we propose a dynamic data augmentation method based on the generator outputs.
Specifically, gor each training mini-batch, we employ the associated generator outputs to generate synthetic missing data.
Subsequently, we establish a reconstruction loss function for these augmented incomplete data,
which mandates that the imputed outcomes from synthetic missing data remain consistent with the raw observations.
We incorporate this reconstruction loss as a regularization term into the optimization objective of the original model.
In this way, the proposed method does not need to alter the training process of the original model and works for various model structures and data types.
We refer to the proposed method as Missingness Augmentation (MisA).

MisA offers several advantages, as outlined below:
(1) As shown in Section \ref{MAug}, MisA uses the outputs of the generator and artificial masks to yield augmented incomplete data,
and therefore does not rely on statistical correlation between attributes,
making it suitable for highly structured homogeneous data such as images, as well as heterogeneous tabular data such as UCI \cite{Dua2017}.
(2) MisA is easy to implement and only requires a simple mean-squared reconstruction loss as regularization. Despite its simplicity, it leads to considerable performance improvements.
(3) MisA is effective for a variety of deep generative imputation models, leading to better convergence of the original training loss curve (see Figure \ref{LossCur}).
When we visualize the augmented data and we can find that  they gradually approach the distribution of the real data as the training proceeds (see Figure \ref{DisAug}).

In summary, the main contributions of the paper are as follows:
\begin{itemize}
	\item
     We propose a novel idea that the imputed values should accurately reflect the raw observations, a concept that has not been emphasized in previous work.
    \item
    Building on this idea, we introduce a new approach, called Missingness Augmentation (MisA),
    which is simple, efficient, and applicable to a wide range of deep generative imputation models
    for both highly structured homogeneous data such as images and heterogeneous tabular data such as the UCI dataset.
	\item
    We evaluate MisA using benchmark deep generative imputation models and demonstrate its efficacy through significant performance improvements on popular datasets.
    These results indicate the potential of the approach to benefit practical applications in various domains.
\end{itemize}

\section{Related Works}
\subsection{Generative Imputation Models}
Variational autoencoders (VAEs) \cite{Vincent2010}  are efficient
and accurate in capturing the latent structure of large amounts of complex high-dimensional data.
Two imputation methods,  HI-VAE \cite{Nazabal2020} and MIVAE \cite{Mattei2019}, were developed to maximize the evidence lower bound on incomplete data
under the assumption of MCAR and MAR missing mechanisms, respectively.
Meanwhile, HI-VAE provides useuful  guidelines for dealing with heterogeneous tabular data.
Richardson et al. \cite{Richardson2020} developed a generative imputation model that leverages a normalizing flow as the underlying density estimator.
On the other hand, there is an exploding growth in studies of generative adversarial nets (GANs) \cite{goodfellow2014generative} for missing data imputation.
Pathak et al. \cite{Pathak2016} presented an encoder that effectively extracts context features from observed images,
and a decoder that produces the missing regions according to features under the joint supervision of adversarial loss and reconstruction loss.
The seminal paper of Yoon et al. \cite{Yoon2018} fit the implicit relationship between the observed and complete data,
in which the discriminator takes an additional hint matrix as input to help distinguish real complete data from imputed data.
Li et al. \cite{Li2019} adopted two sets of GANs to decouple the mask from corresponding incomplete data,
which achieved state-of-the-art performance under the MCAR assumption.
Zhang et al. \cite{Zhang2021} considered an end-to-end generative adversarial model to impute the missing values in a multivariate time series.
Recently, Lee et al. \cite{Lee2022} utilized cycle-consistent imputation adversarial networks to discover the underlying distribution of missing patterns closely under some relaxations.

Current state-of-the-art models typically involve multiple modules,
which can be challenging for limited computing resources.
Additionally, when training data is limited, imputation quality can quickly deteriorate.
In this paper, we propose a "plug-and-play" regularization approach that can improve the performance of existing generative missing data imputation models with little computational overhead.

\subsection{ Data Augmentation}
Data augmentation is a set of techniques to  increase the amount of training samples by generating virtual data from existing ones.
In the computer vision field, digital images can be easily enriched through rotation and cropping \cite{He2016},
as well as some elaborate augmentations such as CutOut \cite{Devires2017} and CutMix \cite{Yun2019}.
In the natural language processing field, replacing some words by their synonyms \cite{Zhang2015}
or masking part of the sentence \cite{Devlin2019} preserves most of inherent semantics.
In addition, some sentence-level data augmentations \cite{Kuo2020, Li2021}
perform the transformation in a learned feature space rather than only conduct augmentation in the input space.
However, these augmentation techniques rely heavily on the statistical dependencies of pixels or words.
But practical tabular data in medical, financial and biological fields do not contain such spatial or sequential dependencies between the attributes.

Only a limited number of data augmentation techniques can be effectively applied to tabular data.
Two examples are denoising \cite{Vincent2008}, which introduces noise into the data and then recovers the original values,
and masking \cite{Pathak2016,Yoon2020}, which adds a mask to the raw sample and then imputes the missing parts or estimates the mask vector.
However, these techniques were primarily developed for complete data and may struggle to capture the characteristics of incomplete data,
leading to poor results in the presence of high missingness.

As a consequence, there is an urgent need to develop novel data augmentation techniques that can effectively address incomplete data imputation tasks on various datasets.

\section{Proposed Approach}

\subsection{Preliminaries}
Let $ \mathbf{\chi}  \in \mathbb{R}^d $ denote an incomplete dataset.
For each sample $ \mathbf{x}_m \in \mathbf{\chi}  $, there exists a corresponding binary mask vector  $ \mathbf{m}= \{0,1\}^d $,
where $ m_i = 1 $ if the $i$-th feature $ x_{m,i} $ is observed, and $ m_i = 0 $ if $ x_{m,i} $ is missing.
During training, missing attributes are usually filled by zero values or noise from a specific distribution.

Let $ G_{\theta}(\cdot) $ be any deep generative imputation model,
which takes the incomplete sample $ \mathbf{x}_m $ and the mask $ \mathbf{m} $ as input
and then yield the imputed result $ \mathbf{x}_G  $  by
\begin{align}
	\mathbf{x}_G = (1 - \mathbf{m}) \odot G_{\theta}(\mathbf{x}_m, \mathbf{m}) + \mathbf{m} \odot \mathbf{x}_m .
\end{align}
Without loss of generality, let $ \mathcal{L}_{\textrm{ori}}(\mathbf{x}_G, \mathbf{x}_m, \mathbf{m}) $ denote the original composite loss of the model $ G_{\theta}(\cdot) $.

The majority of prior studies have concentrated on enhancing imputation outcomes by designing complex network frameworks.
Nonetheless, it has been observed that in many cases, the advancements derived from these intricate models are marginal.
We believe that a reliable imputation model should not only be able to infer missing values from observed data, but also produce imputed values that accurately reflect the raw observations.
In light of this, we are to propose a dynamic data augmentation technique that can be seamlessly integrated into various deep generative imputation models.

\subsection{Missingness Augmentation}
\label{MAug}

\begin{figure*}[htb]
	\centering
	\scalebox{0.36}{\includegraphics{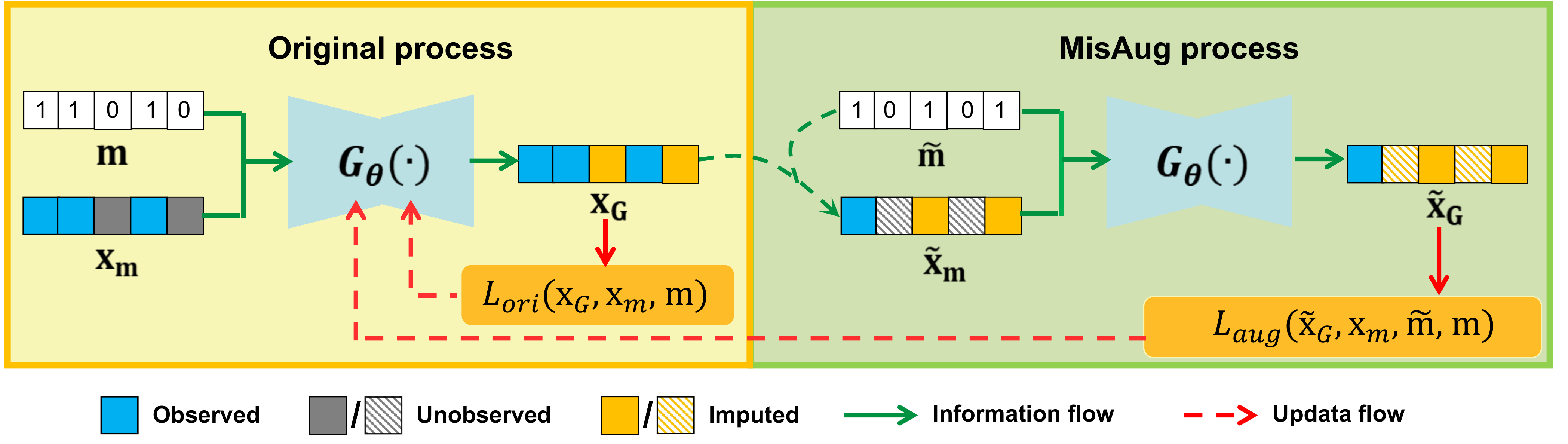}}
	\caption{The framework of Missingness Augmentation. The model is end-to-end trainable with an additional reconstruction loss $ \mathcal{L}_{\mathrm{aug}} $.}
	\label{framework}
\end{figure*}

For any incomplete sample $ \mathbf{x}_m $ and mask $\mathbf{m}$,
define  an artificial binary mask  $ \tilde{\mathbf{m}} \in \{0,1\}^d $ for the augmented samples such that
\begin{equation}
\label{eqt_p}
\begin{cases}
&\mathbf{P}(\tilde{m}^i = 0) =  \displaystyle\frac{1}{d}\sum^d_{i=1} (1 - m_i), \\
&\mathbf{P}(\tilde{m}^i = 1) =  \displaystyle\frac{1}{d}\sum^d_{i=1} m_i,
\end{cases}
\end{equation}
where $ \mathbf{P} $ denotes the probability measure.
Note that the artificial missing rate defined in \eqref{eqt_p} is consistent with the proportion of the observed attributes in the raw sample.
Such definition is inline with the starting point that the generated values should be capable of inferring the raw observations.

With the output of the generator and the artificial missing rate $ \tilde{\mathbf{m}} $,  we can build the augmented sample as follows:
\begin{align}
\label{AugData}
	\tilde{\mathbf{x}}_m = \tilde{\mathbf{m}} \odot \mathbf{x}_G + (1 - \tilde{\mathbf{m}}) \odot \mathbf{z},
\end{align}
where $ \mathbf{z} $ is a noise vector, which should be consistent with the filled values in the raw incomplete data.

As depicted by the left part of Figure \ref{framework},
in each epoch the generator $ G_{\theta}(\cdot) $ takes the augmented sample $\tilde{\mathbf{x}}_m$ and the artificial mask $\tilde{\mathbf{m}} $ as the input,
and yields the extra imputed result
\begin{align}
	\tilde{\mathbf{x}}_G = (1 - \tilde{\mathbf{m}}) \odot G_{\theta}(\tilde{\mathbf{x}}_m, \tilde{\mathbf{m}}) + \tilde{\mathbf{m}} \odot \tilde{\mathbf{x}}_m .
\end{align}
We require the imputed values of the augmented sample consistent with the raw observations.
To achieve this goal, we consider the following reconstruction loss for augmented samples:
\begin{align}
\label{penalty}
\begin{split}
\mathcal{L}_{\mathrm{aug}} (\tilde{\mathbf{x}}_G , \mathbf{x}_m, \tilde{\mathbf{m}}, \mathbf{m})
 = \|\mathbf{m} \odot (1 - \tilde{\mathbf{m}}) \odot (\tilde{\mathbf{x}}_G - \mathbf{x}_m) \|_2^2.
\end{split}
\end{align}
This reconstruction loss is then embedded as a regularization term into the optimization objective of the original model. As a result, the final optimization objective is formulated as follows:
\begin{align}
\label{hybridloss}
   \min_{G} \mathbb{E}_{\mathbf{x}_m \in \mathbf{\chi} } \{\mathcal{L}_{\textrm{ori}}(\mathbf{x}_G, \mathbf{x}_m, \mathbf{m})
 +  \alpha \mathcal{L}_{\mathrm{aug}} (\tilde{\mathbf{x}}_G , \mathbf{x}_m, \tilde{\mathbf{m}}, \mathbf{m})\},
\end{align}
where $ \alpha > 0 $ is a regularization hyperparameter,
which should be chosen such that the first and second terms are of the same order of magnitude at the beginning of training.
By minimizing the objective \eqref{hybridloss}, the model not only learns to impute missing values from observations but also infers observations from imputed values.

\begin{rmrk}
It should be noted that the augmented samples in our approach are not directly generated from the raw data,
but dynamically generated from the output of a mini-batch at the beginning of each training epoch.
As a result, our approach can be regarded as a type of intermediate-data-augmentation technique.
In Section \ref{REff}, Figure \ref{DisAug} demonstrates that the augmented samples gradually approach the distribution of the raw data as the training iterations progress.
\end{rmrk}

Algorithm \ref{alg} outlines the training procedure with Missingness Augmentation.

\begin{algorithm}[htb]
  \caption{Training Procedure with Missingness Augmentation}
  \label{alg}
  \begin{algorithmic}[1]
    \Require
      Incomplete data set $\mathcal{\chi}$,
      imputation model $G_{\theta}(\cdot)$,
      and regularization coefficient $ \alpha>0 $;
     \For{each mini-batch}
     \For{ each $ \mathbf{x}_m $ in the min-batch}
         \State Construct the artificial mask vector $\tilde{\mathbf{m}}$ via \eqref{eqt_p};
         \State Generate the augmented sample $\tilde{\mathbf{x}}_m$ via \eqref{AugData};
    \EndFor
	\State Update the model $G_{\theta}(\cdot)$ by minimizing the modified objective \eqref{hybridloss} using the mini-batch and the corresponding augmented samples.
	\EndFor
  \end{algorithmic}
\end{algorithm}

\section{Experiments}
\subsection{Experimental Setup}
\subsubsection{Datasets}
In this section, we evaluate the performance of our approach on the MNIST dataset \cite{LeCun2010},  the CIFAR-10 dataset \cite{Krizhevsky2014}, the CelebA dataset \cite{Liu2015} and eight datasets from the UCI repository \cite{Dua2017}.
The details of the UCI datasets are listed in Table \ref{UCI}.

\begin{table}[!htb]
\center
\caption{The basic properties of the UCI datasets.}
\label{UCI}
\scalebox{0.85}{
\begin{tabular}{ccccc}
\hline
&	Samples	&	Numerical variables	&	Categorial variables	&	Number of classes     \\
\cline{2-5}							
Abalone            &4177        &7                     &1                       &3                   \\
Avila              &20867       &10                    &0                       &12                  \\
Ionosphere         &351         &32                    &2                       &2                   \\	
News               &39797       &35                    &23                      &2				    \\
Pendigits          &10992       &16                    &0                       &10                  \\
Sonar              &208         &60                    &0                       &2				    \\
Wine               &178         &13                    &0                       &3				    \\
WineQuality        &4898        &11                    &0                       &7                   \\
\hline
\end{tabular}}
\end{table}

Since no dataset contains missing values initially,
for a given missing rate,
we remove the features of all data completely at random to formulate an incomplete dataset.
Moreover, each variable is scaled to the interval  $ [0,1] $.

\subsubsection{Baseline imputation models}
We use six deep generative imputation models to illustrate the effectiveness the proposed method.
\begin{itemize}
  \item\textbf{DAE} \cite{Gondara2018}:
  DAE is an imputation method based on overcomplete deep denoising autoencoders.
  It can deal with multiple data missing cases.
    \item \textbf{DAEMA} \cite{Tihon2021}: DAEMA is a benchmark imputation model using the denoising autoencoder architecture with an attention mechanism.
  \item \textbf{GAIN} \cite{Yoon2018}:
  GAIN is a missing data imputation method based on generative adversarial nets.
  It introduces a hint vector in the adversarial imputation process to improve the performance quality.
  \item \textbf{MCFlow} \cite{Richardson2020}:
  MCFlow is a method that leverages normalizing flow generative models and Monte Carlo sampling  for incomplete data imputation.
  \item \textbf{VAEAC} \cite{Nazabal2020}:
   VAEAC is a recently developed imputation model based on variational autoencoders framework.
  \item \textbf{MisGAN} \cite{Li2019}:
  MisGAN is also a GAN based imputation method, which learns a complete data generator along with a mask generator that models the missing data distribution.
  It has achieved the state-of-the-art performance for data imputation on the CelebA dataset.
\end{itemize}
For all imputation models, we adopt the suggested settings of the original literature
and take 5-cross validation in the experiments.
Each experiment is repeated 10 times and the average performance is reported.

\subsubsection{Hyperparameters}
Unless stated otherwise,
the performance of each model is evaluated at a completely random 50\% missing rate.
The regularization coefficient $\alpha $ is set to make the first and second terms in the optimization objective \eqref{hybridloss} are of the same scale at the beginning of training.

\begin{table}[htb]
	\center
	\caption{The default choice of regularization coefficient $ \alpha $ for different imputation models}
	\label{tableA}
	\scalebox{0.8}{
		\begin{tabular}{ccccccc}
			\hline
			            &DAE \cite{Gondara2018}   &DAEMA \cite{Tihon2021}  &GAIN \cite{Yoon2018}  &MCFlow \cite{Richardson2020} &MisGAN \cite{Li2019}  &VAEAC \cite{Nazabal2020}  \\
			\cline{2-7}
			$ \alpha  $  &5                       &50                     &100                  &1                           &1                    &50 \\
			\hline
	\end{tabular}}
\end{table}

\subsection{ Imputation Accuracy on UCI Datasets}
We use the Root Mean Square Error (RMSE),
which computes the root mean square error of the imputed missing values against the
ground truth,
as the metric to evaluate the accuracy of the imputed results on UCI datasets.
The smaller the RMSE, the more accurate the result.

The experimental results presented in Table \ref{UCI_RMSE} provide compelling evidence of the efficacy of our proposed augmentation approach on tabular datasets.
Specifically, our approach outperforms the corresponding baseline model significantly in the majority of cases, underscoring its effectiveness.

\begin{table}[!hbt]
\center
\caption{Comparison of RMSE between the  baseline models and their corresponding augmented versions  ($ '+' $) on the UCI datasets under a 50\% missing rate.}
\label{UCI_RMSE}
\scalebox{0.56}
{\begin{tabular}{ccccccccc}
\hline
                           &Abalone                     &Avila                       &Ionosphere                 &News                         &Pendigits                  &Sonar              &Wine                  &WineQuality\\
\cline{2-9}
DAE \cite{Gondara2018}      &0.1424$\pm$0.0309           &0.2325$\pm$0.0844           &0.2476$\pm$0.0049          &\textbf{0.2360}$\pm$0.0032   &0.2385$\pm$0.0016          &0.2028$\pm$0.0066  &0.2110$\pm$0.0068     &0.1450$\pm$0.0115\\
DAE+                       &\textbf{0.1292}$\pm$0.0182  &\textbf{0.1262}$\pm$0.0636  &\textbf{0.2410}$\pm$0.0048 &0.2401$\pm$0.0162            &\textbf{0.2285}$\pm$0.0016 &\textbf{0.1919}$\pm$0.0040  &\textbf{0.2015}$\pm$0.0112     &\textbf{0.1201}$\pm$0.0066\\
\hline
DAEMA \cite{Tihon2021}       &\textbf{0.1189}$\pm$.0263  &0.1167$\pm$.0434               &0.2742$\pm$.0084               &0.2451$\pm$.0058               &0.2219$\pm$.0084              &0.3242$\pm$.0082               &0.2078$\pm$.0086              &0.1218$\pm$.0261       \\
DAEMA+  &0.1192$\pm$.0193              &\textbf{0.1012}$\pm$.0376   &\textbf{0.2647}$\pm$.0087   &\textbf{0.2441}$\pm$.0047   &\textbf{0.1987}$\pm$.0077  &\textbf{0.2358}$\pm$.0053   &\textbf{0.1988}$\pm$.0095  &\textbf{0.1207}$\pm$.0292 \\
\hline
GAIN \cite{Yoon2018}               &0.1597$\pm$0.0476  &0.1224$\pm$0.0343  &0.2941$\pm$0.0082   &0.2491$\pm$0.0071    &0.2323$\pm$0.0092  &0.3035$\pm$0.0075  &0.2476$\pm$0.0093     &0.1397$\pm$0.0185\\
GAIN+					          &\textbf{0.1067}$\pm$0.0445  &\textbf{0.1020}$\pm$0.0115  &\textbf{0.2509}$\pm$0.0069  &\textbf{0.2173}$\pm$0.0104    &\textbf{0.1743}$\pm$0.0025	&\textbf{0.2480}$\pm$0.0048  &\textbf{0.2170}$\pm$0.0119     &\textbf{0.1224}$\pm$0.0125\\
\hline
MCFlow \cite{Richardson2020}       &0.0880$\pm$0.0008  &0.0962$\pm$0.0015  &0.2311$\pm$0.0045   &\textbf{0.2405}$\pm$0.0019    &0.1356$\pm$0.0156  &0.1613$\pm$0.0020  &0.2057$\pm$0.0070     &0.1143$\pm$0.0012\\
MCFlow+					          &\textbf{0.0871}$\pm$0.0004  &\textbf{0.0922}$\pm$0.0023  &\textbf{0.2272}$\pm$0.0054   &0.2412$\pm$0.0005    &\textbf{0.1202}$\pm$0.0225	&\textbf{0.1587}$\pm$0.0020  &\textbf{0.1991}$\pm$0.0090     &\textbf{0.1132}$\pm$0.0020\\
\hline
VAEAC \cite{Nazabal2020}           &0.1231$\pm$0.0276  &0.1158$\pm$0.0065  &0.2419$\pm$0.0047  &0.2314$\pm$0.0013    &0.2124$\pm$0.0017
&0.2309$\pm$0.0027  &0.1995$\pm$0.0066     &0.1445$\pm$0.0077\\
VAEAC+	                          &\textbf{0.0919}$\pm$0.0152  &\textbf{0.1043}$\pm$0.0088  &\textbf{0.2282}$\pm$0.0048   &\textbf{0.2263}$\pm$0.0029    &\textbf{0.2005}$\pm$0.0019   &\textbf{0.1923}$\pm$0.0038  &\textbf{0.1899}$\pm$0.0066     &\textbf{0.1204}$\pm$0.0063\\
\hline
\end{tabular}}
\end{table}

\begin{table}[!hbt]
\center
\caption{Comparison of RMSE between the  baseline models and their corresponding augmented versions  ($ '+' $)  with various missing rates.}
\label{VariMiss}
\scalebox{0.7}
{\begin{tabular}{ccccccccccccc}
\hline
\multirow{2}*                  &\multicolumn{4}{c}{Abalone}                                     &\multicolumn{4}{c}{Ionosphere}                            &\multicolumn{4}{c}{Wine}                                               \\
\cline{2-13}
                               & 20\%             & 40\%         & 60\%         &  80\%         & 20\%       & 40\%         & 60\%         &  80\%         & 20\%       & 40\%         & 60\%         &  80\%            \\
\hline
DAE \cite{Gondara2018}          &0.1201            &0.1354        &0.1537        &0.1803         &0.2145      &0.2321        &0.2587        &0.2932         &0.1734      &0.1983        &0.2358        &0.2872 \\
DAE+                           &0.1111            &0.1203        &0.1348        &0.1579         &0.2119      &0.2252        &0.2480        &0.2865         &0.1697      &0.1870        &0.2133        &0.2534 \\
\hline
DAEMA \cite{Tihon2021}          &0.1002            &0.1097        &0.1339        &0.1790         &0.2164      &0.2457        &0.2972        &0.3904         &0.1877      &0.1978        &0.2436        &0.3320      \\
DAEMA+                         &0.1012            &0.1085        &0.1301        &0.1719         &0.2127      &0.2316        &0.2716        &0.3489         &0.1790      &0.1889        &0.2320        &0.3167  \\
\hline
GAIN \cite{Yoon2018}            &0.1220            &0.1446        &0.1788        &0.2371         &0.2247      &0.2663        &0.3293        &0.4367         &0.1892      &0.2242        &0.2773        &0.3677              \\
GAIN+                          &0.0988            &0.1092        &0.1323        &0.1662         &0.2201      &0.2358        &0.2968        &0.3887         &0.1719      &0.1999        &0.2413        &0.3200          \\
\hline
MCFlow \cite{Richardson2020}    &0.0778            &0.0820        &0.0991        &0.1369         &0.1673      &0.1891        &0.2362        &0.3236         &0.1673      &0.1891        &0.2362        &0.3236           \\
MCFlow+                        &0.0674            &0.0762        &0.0952        &0.1304         &0.2017      &0.2126        &0.2529        &0.3533         &0.1790      &0.1889        &0.2320        &0.3167  \\
\hline
VAEAC \cite{Nazabal2020}        &0.1112            &0.1184        &0.1337        &0.1508         &0.2204      &0.2390        &0.2534        &0.2714         &0.1788      &0.1872        &0.2251        &0.2692     \\
VAEAC+                         &0.1052            &0.1077        &0.1155        &0.1256         &0.2182      &0.2253        &0.2315        &0.2538         &0.1709      &0.1754        &0.2023        &0.2257        \\
\hline
\end{tabular}}
\end{table}

We further use the Abalone, Ionospher and Wine to evaluate the imputation performance under various missing rates.
As shown in  Table \ref{VariMiss},
the augmented version consistently outperforms its original model.
Especially, the advantage becomes more obvious when the missing rate becomes higher.

\subsection{Performance for Incomplete Image Data}
In this section we evaluate the performance of MisA  on various image datasets.
To fully assess the quality,
we use respectively the Peak Signal-to-Noise Ratio (PSNR),
Structural Similarity Index (SSIM) and  Fr\'echet Inception Distance (FID)  to measure the quality of the imputed results.
The higher the PSNR or SSIM, the better the imputation quality.
On the contrary, the lower the FID, the better the quality.

\begin{figure}[!hbt]
\centering
\subfloat[Incomplete images, where gray pixels indicate missing values]{\label{fig:mdleft}{\includegraphics[width=0.3\textwidth]{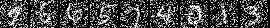}}}\hfill
\subfloat[GAIN \cite{Yoon2018},  GAIN+ ]{\label{fig:mdleft}{\includegraphics[width=0.48\textwidth]{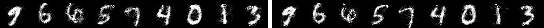}}}\hfill
\subfloat[MCFlow \cite{Richardson2020},  MCFlow+]{\label{fig:mdright}{\includegraphics[width=0.48\textwidth]{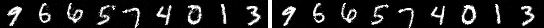}}}\hfill
\subfloat[MisGAN \cite{Li2019},  MisGAN+ ]{\label{fig:mdleft}{\includegraphics[width=0.48\textwidth]{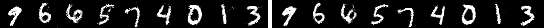}}}\hfill
\caption{Visualization of the imputed results of baseline models (left) and their corresponding augmented versions (right, indicated with $ '+' $) on the MNIST dataset under a 50\% missing rate.}
\label{MNIST}
\end{figure}

As shown in Table \ref{Image_PSNR},
all augmented versions have better PSNR, SSIM and FID scores than the corresponding baselines.
Further, it can be observed from Figure \ref{MNIST} that
the imputed images based on MisA  are more cohesive and smoother.

\begin{table}[hbt]
\center
\caption{Comparison between the  baseline models and their corresponding augmented versions  ($ '+' $) on the MNIST and CIFAR-10 datasets under a 50\% missing rate.}
\label{Image_PSNR}
\scalebox{0.85}
{\begin{tabular}{ccccccc}
\hline
\multirow{2}*       &\multicolumn{3}{c}{MNIST}                                                     &\multicolumn{3}{c}{CIFAR-10}                                               \\
\cline{2-7}
                            &PSNR$\uparrow$       &SSIM$\uparrow$          &FID$\downarrow$          &PSNR$\uparrow$           &SSIM$\uparrow$       &FID$\downarrow$             \\
\hline
DAEMA \cite{Tihon2021}       &24.94                       &0.892                  &0.144                    &27.81                    &0.945                           &0.127           \\
DAEMA+                              &\textbf{26.06}   &\textbf{0.930}      &\textbf{0.137}     &\textbf{28.35}   &\textbf{0.952 }    &\textbf{0.118} \\
\hline
GAIN \cite{Yoon2018}         &22.02                       &0.864                  &0.163                    &25.17                    &0.847                           &0.147              \\
GAIN+                               &\textbf{23.01}     &\textbf{0.901}         &\textbf{0.144 }     &\textbf{25.66}      &\textbf{0.853}              &\textbf{0.132}           \\
\hline
MCFlow \cite{Richardson2020} &27.05                       &0.967                  &0.100                    &30.30                    &0.956                           &0.083           \\
MCFlow+                               &\textbf{28.33 }       &\textbf{0.975}    &\textbf{0.098}      &\textbf{31.73 }     &\textbf{0.974}          &\textbf{0.079} \\
\hline
MisGAN \cite{Li2019}         &25.56                       &0.924                  &0.123                    &28.15                    &0.933                           &0.130           \\
MisGAN+                           &\textbf{26.61}      &\textbf{0.963}   &\textbf{0.119}       &\textbf{29.24}           &\textbf{0.963}    &\textbf{0.121} \\
\hline
\end{tabular}}
\end{table}

\begin{figure}[hbt]
\centering
\subfloat[Real images]{\label{fig:mdleft}{\includegraphics[width=0.24\textwidth]{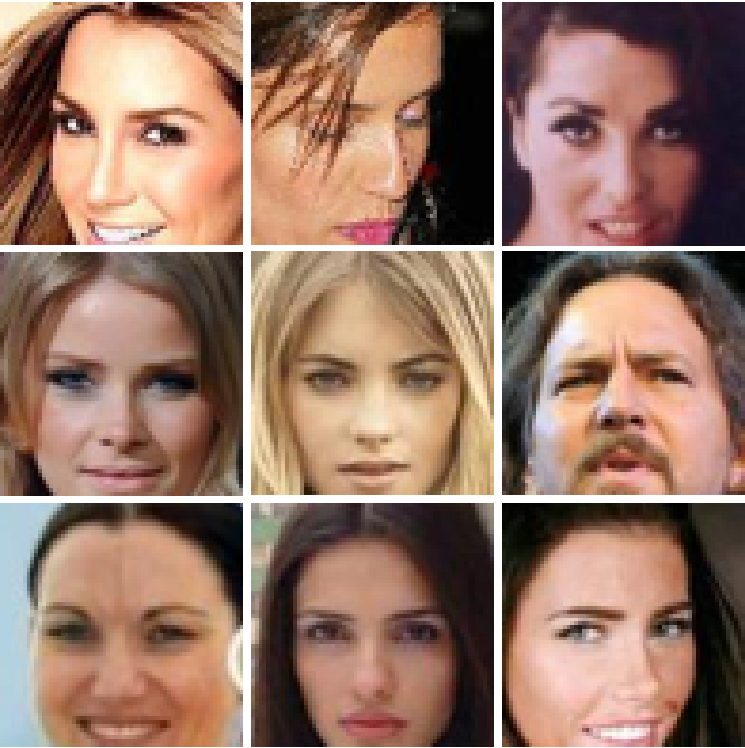}}}\hfill
\subfloat[Incomplete images]{\label{fig:mdright}{\includegraphics[width=0.24\textwidth]{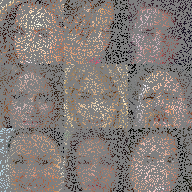}}}\hfill
\subfloat[MisGAN\cite{Li2019}]{\label{fig:mdright}{\includegraphics[width=0.24\textwidth]{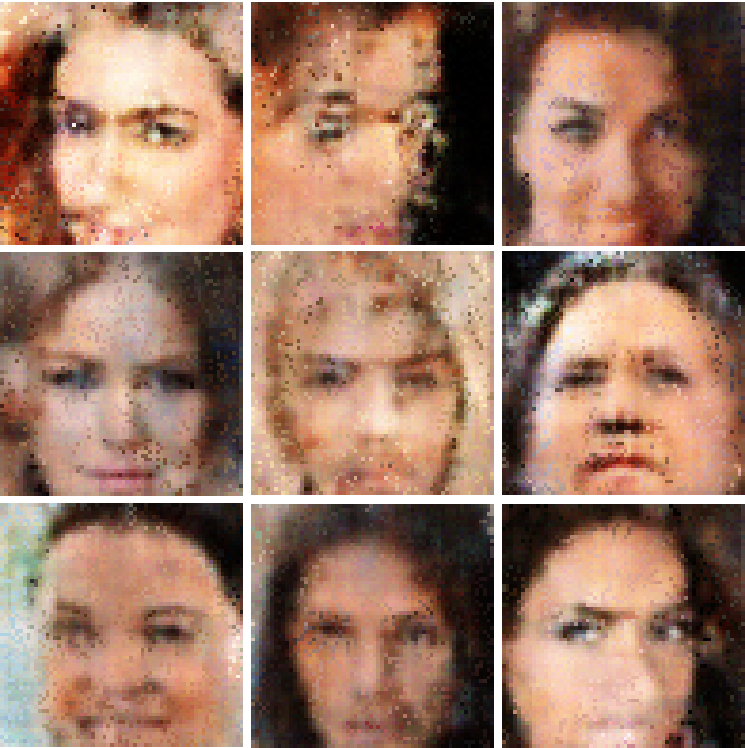}}}\hfill
\subfloat[MisGAN+ ]{\label{fig:mdleft}{\includegraphics[width=0.24\textwidth]{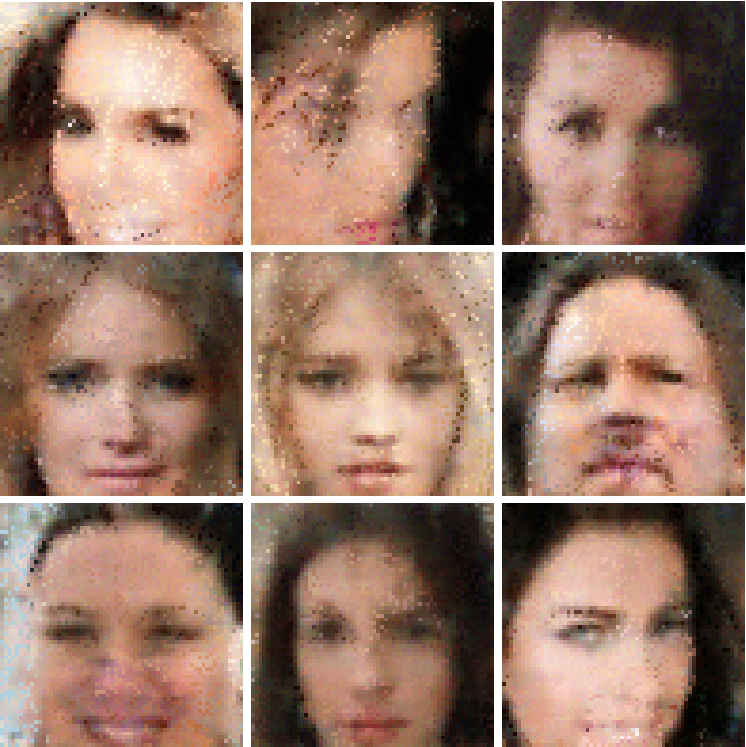}}}
\caption{Visualization of the imputed results by MisGAN and its augmented version on the CelebA dataset under a 80\% missing rate.}
\label{CelebA}
\end{figure}

\smallskip
Compared to the MNIST dataset, the CelebA dataset consists of images with higher complexity and resolution,
making it more challenging to impute accurately.
In this study, we use MisGAN \cite{Li2019}, which has achieved state-of-the-art performance on the CelebA dataset,
as an example to evaluate the performance of our approach.
For each CelebA image, we randomly dropped 80\% of the pixels.
Figure \ref{CelebA} (c) illustrates that MisGAN fails to satisfactorily restore the texture of the missing parts in some areas, leading to unpleasant artifacts.
This problem arises partly because of the uncertainty introduced by the masks used in MisGAN.
Our approach, however, can alleviate this issue and thus yield better results.

The qualitative and quantitative analyses in this subsection demonstrate that
the proposed augmentation method can enhance the performance of generative imputation models on various image datasets.
These findings suggest that our approach can be a useful tool for addressing missing information in complex images.

\subsection{Post-imputation Prediction Accuracy}

\begin{table}[!hbt]
\center
\caption{Post-imputation prediction accuracy of the original models and the corresponding augmented versions under a 50\% missing rate.}
\label{ab_tab2}
\scalebox{0.8}
{\begin{tabular}{cccccc}
\hline
                            &Abalone          &Avila           &Winequality     &MNIST          &CIFAR-10  \\
\cline{2-6}
DAE \cite{Gondara2018}       &0.290            &0.359           &0.420           &0.942          &\textbf{0.434}     \\
DAE+                        &\textbf{0.322}   &\textbf{0.369}  &\textbf{0.432}  &\textbf{0.951} &0.427           \\
\hline
DAEMA \cite{Tihon2021}       &0.348            &0.424           &0.496           &0.965          &0.508     \\
DAEMA+                      &\textbf{0.377}   &\textbf{0.492}  &\textbf{0.510}  &\textbf{0.973} &\textbf{0.521}     \\
\hline
GAIN \cite{Yoon2018}         &0.326            &0.313           &0.467           &0.964          &0.348     \\
GAIN+                       &\textbf{0.346}   &\textbf{0.378}  &\textbf{0.497}  &\textbf{0.971} &\textbf{0.472}     \\
\hline
MCFlow \cite{Richardson2020} &0.355            &0.442           &0.519           &0.967          &0.495              \\
MCFlow+                     &\textbf{0.380}   &\textbf{0.492 } &\textbf{0.536}  &\textbf{0.974} &\textbf{0.516}     \\
\hline
MisGAN \cite{Li2019}         &-                &-               &-               &0.981          &0.547         \\
MisGAN+                     &-                &-               &-               &\textbf{0.987} &\textbf{0.552}     \\
\hline
VAEAC \cite{Nazabal2020}     &0.327            &0.465           &0.500           &0.961          &\textbf{0.364}     \\
VAEAC+                      &\textbf{0.372}   &\textbf{0.478}  &\textbf{0.504}  &\textbf{0.963} &\textbf{0.364}     \\
\hline
\end{tabular}}
\end{table}

It is widely recognized that a good imputation model should not only accurately restore missing data
but also preserve the categorical information of the data.
In this section, we evaluate our approach on both UCI datasets and image datasets in terms of post-imputation prediction accuracy.
For the UCI datasets, we employ a two-layer fully connected network with ReLU activation for classification,
while for the image datasets, we adopt the LeNet architecture \cite{LeCun1989} as used in \cite{Richardson2020}.

Our results, presented in Table \ref{ab_tab2}, demonstrate that the augmented version consistently outperforms the original model in terms of classification accuracy across all cases.
This, combined with the high imputation accuracy shown in the previous two subsections, suggests that our proposed MisA approach is a reliable method for deep generative imputation tasks."

\subsection{CPU Time} \label{Computational Cost}
\begin{figure}[!hbt]
\centering
\subfloat[UCI datasets]{\label{fig:mdleft}{\includegraphics[width=0.48\textwidth]{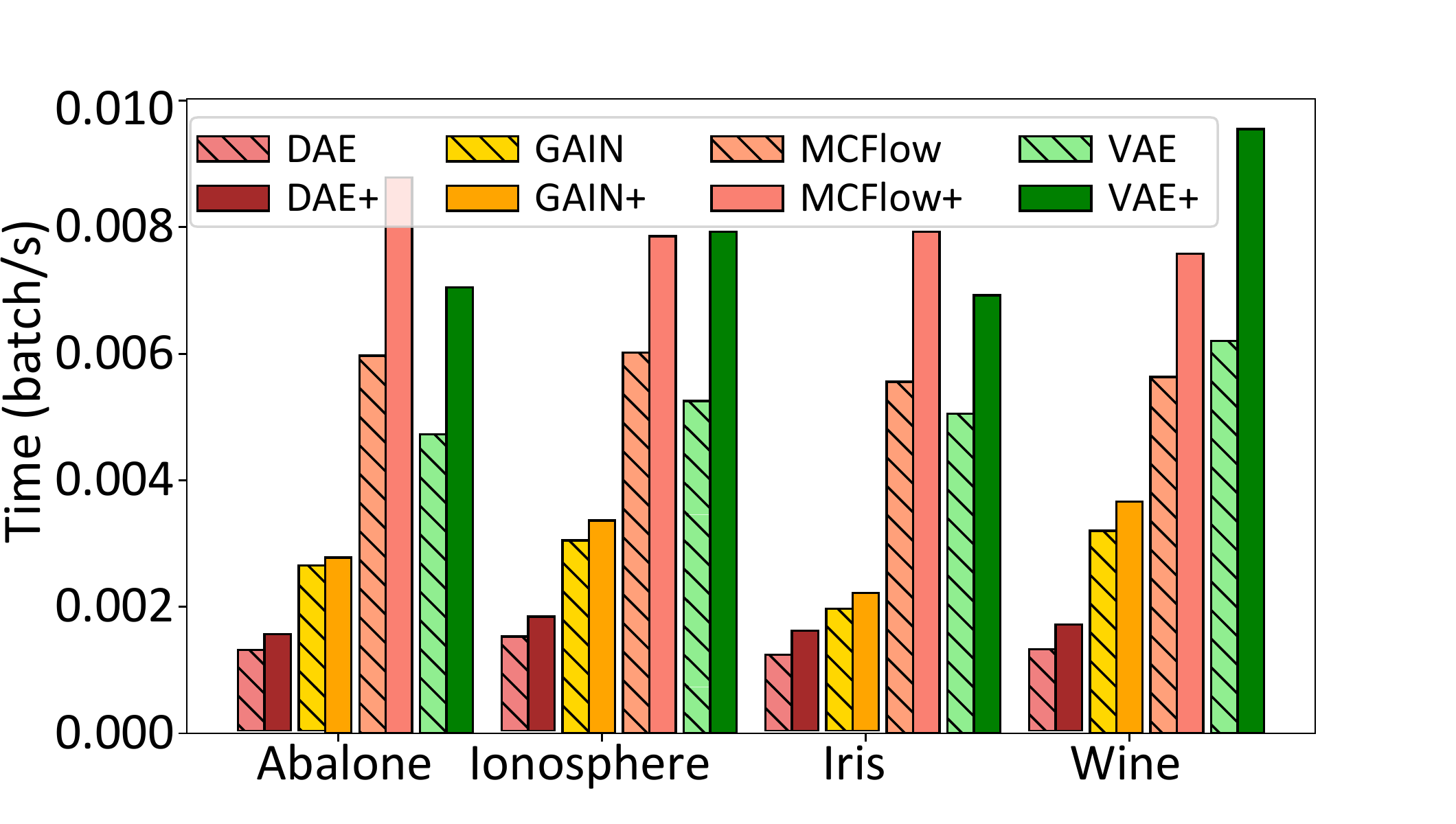}}}
\hfill
\subfloat[MNIST, CIFAR-10 and CelebA datasets]{\label{fig:mdright}{\includegraphics[width=0.48\textwidth]{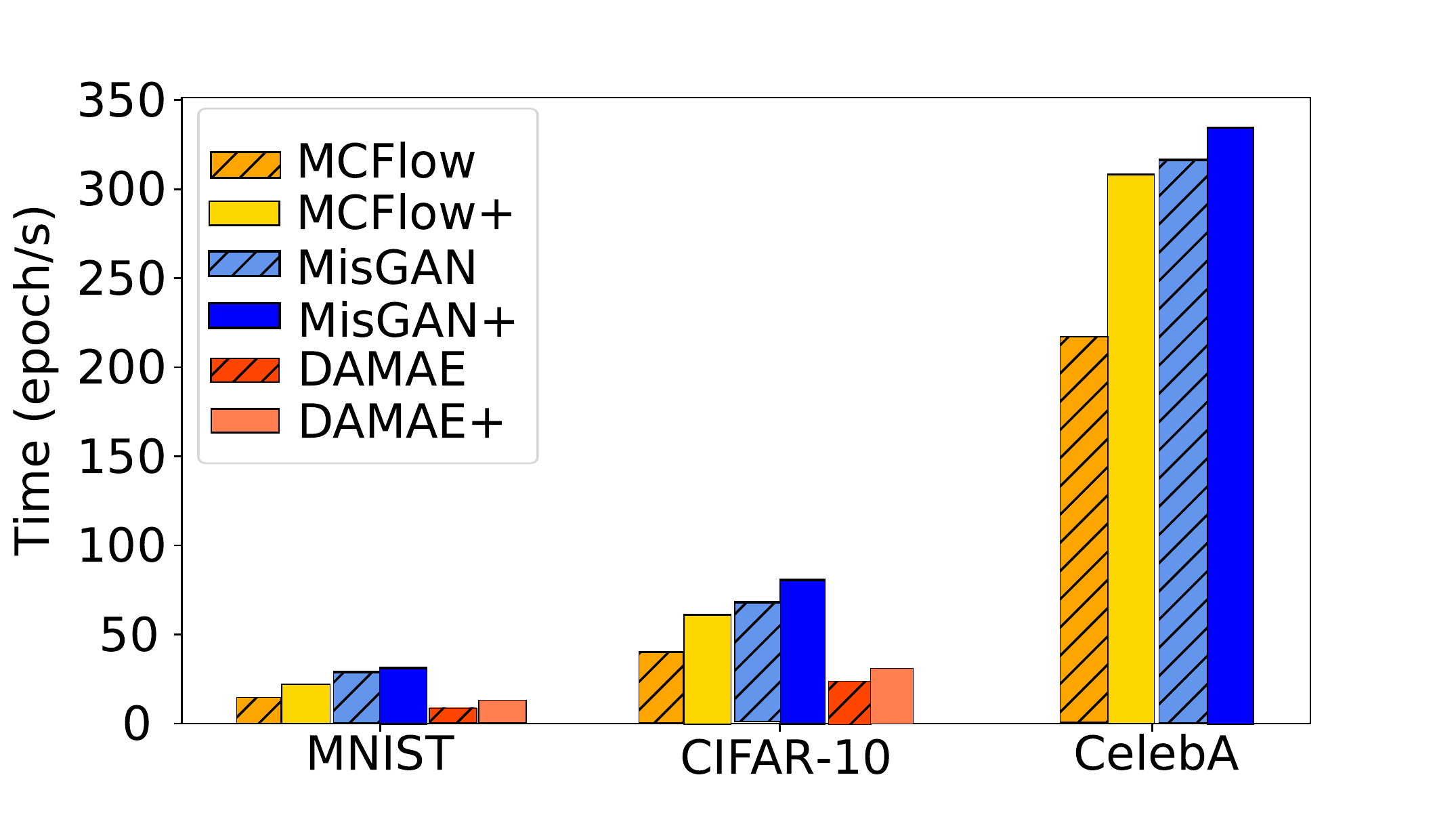}}}
\caption{ Comparisons of training time of imputation models and their corresponding augmented versions.}
\label{TrainingTime}
\end{figure}

The Missingness Augmentation approach is highly efficient and scalable,
and incorporating it into various generative imputation models requires little extra workload.
This advantage is illustrated in Figure \ref{TrainingTime},
which depicts the training time costs of DAE \cite{Gondara2018}, DAEMA \cite{Tihon2021}, GAIN \cite{Yoon2018}, MisGAN \cite{Li2019}, VAEAC \cite{Nazabal2020}, and MCFlow \cite{Richardson2020} along with their corresponding augmented versions.

It is evident that, for DAE, GAIN, and MisGAN, the additional training time accounts for only about 10\%.
Although the time cost for the VAEAC model is slightly higher, it still remains manageable as a special resampling procedure \cite{Nazabal2020} needs to be performed twice during training.
In contrast, the MCFlow model's training time cost is slightly higher, likely due to the inclusion of invertible mapping in the normalizing flow during training.

\subsection{The Reason for Effectiveness}
\label{REff}
The previous subsections have clearly demonstrated the efficacy of MisA in enhancing the performance of deep generative imputation models for both tabular and image data.
This section first examine the distribution of augmented samples at different training stages to understand how MisA affects the data augmentation effect.
We take the imputation model GAIN \cite{Yoon2018} as an instance.

\begin{figure}[hbt]
\centering
\subfloat[Abalone dataset.]{{\includegraphics[width=0.95\textwidth]{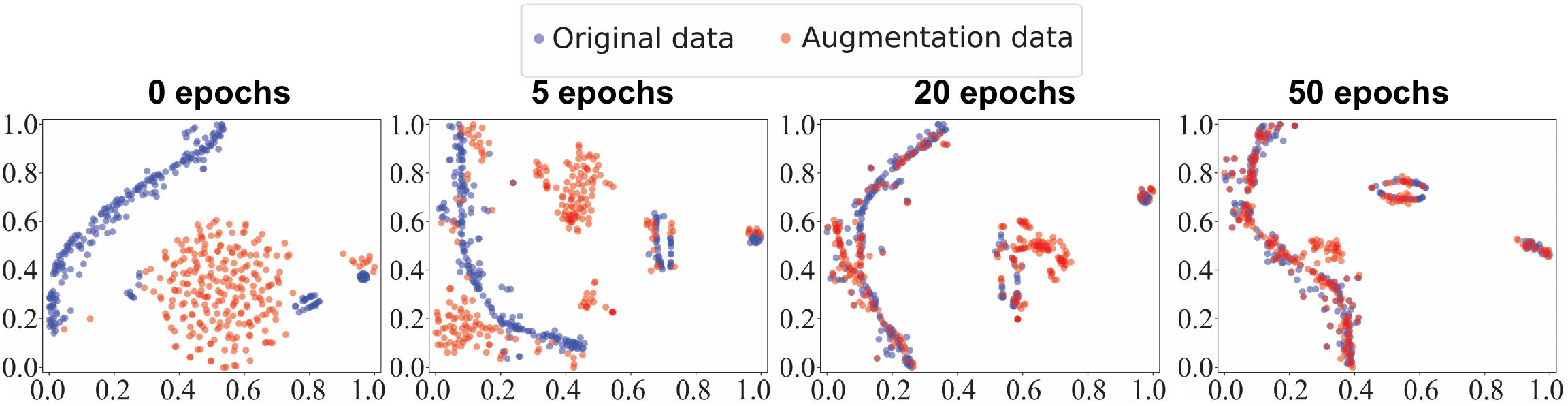}}}\\
\subfloat[WineQuality dataset.]{{\includegraphics[width=0.95\textwidth]{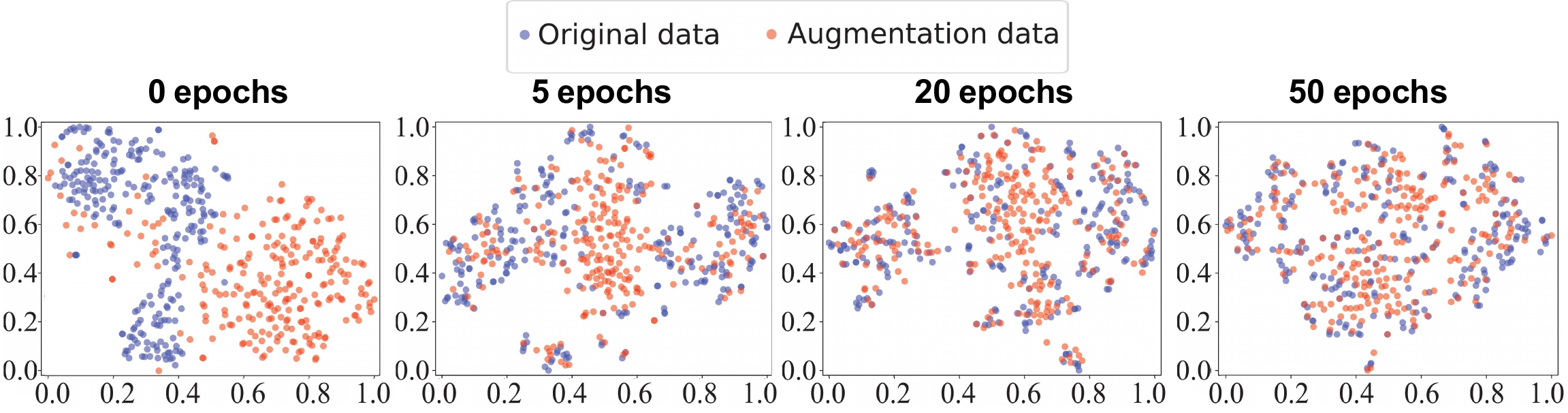}}}\\
\caption{t-SNE \cite{Maa2008} visualization of the real and augmented samples  in different training stages using GAIN \cite{Yoon2018}.
The initial augmented samples deviate from the actual training data, but gradually approach to them as the training progresses.}
\label{DisAug}
\end{figure}

As shown in Figure \ref{DisAug}, during the initial stage, the generator has not been adequately trained,
resulting in augmented samples that deviate from the distribution of actual samples.
However, as training progresses, MisA generates more realistic augmented samples,
thereby helping the generator learn the underlying patterns of the data more effectively,
leading to better imputation performance.

\begin{figure}[!hbt]
\centering
\subfloat[Abalone dataset.]{{\includegraphics[width=0.49\textwidth]{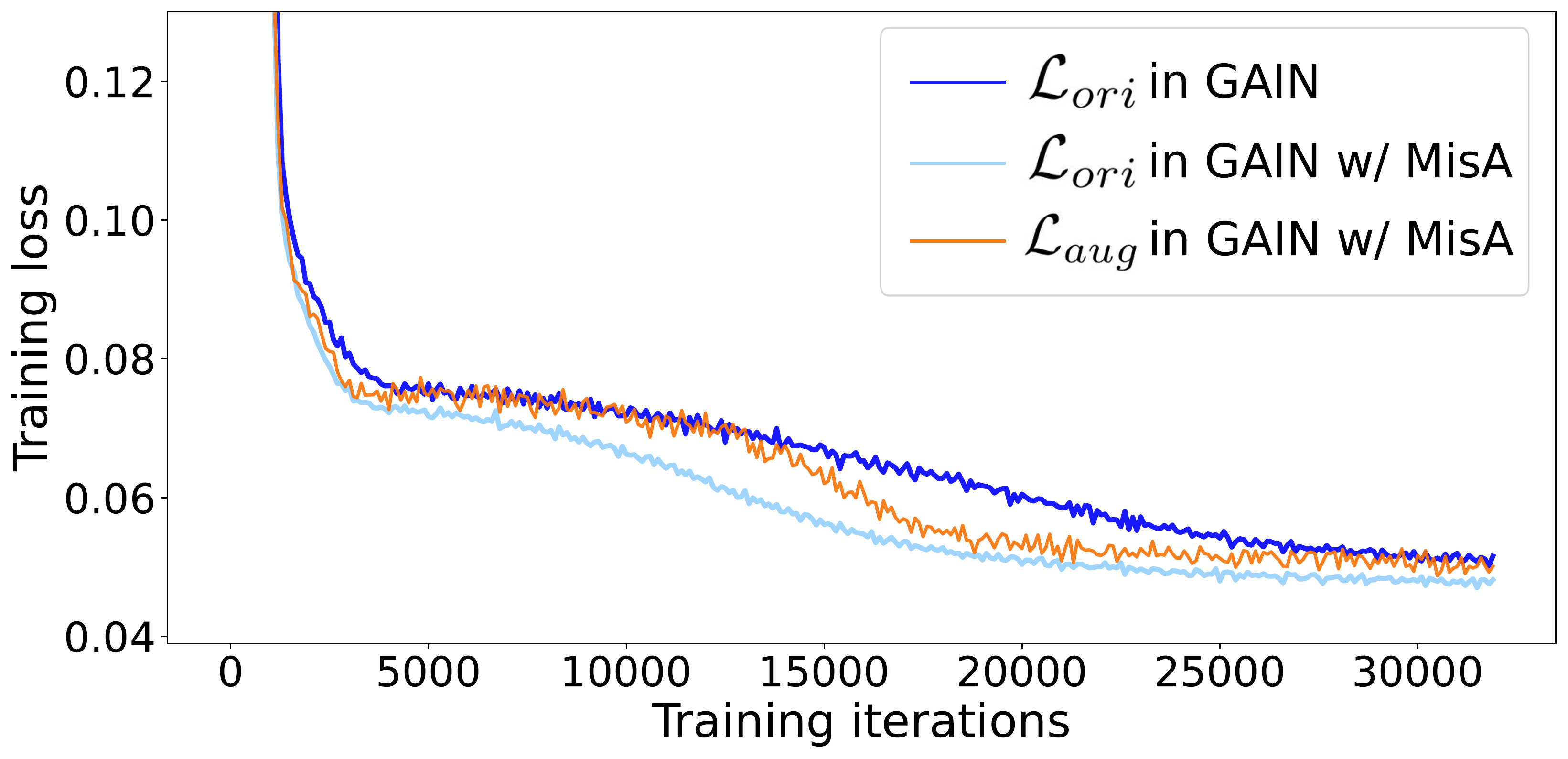}}}
\subfloat[WineQuality dataset.]{{\includegraphics[width=0.5\textwidth]{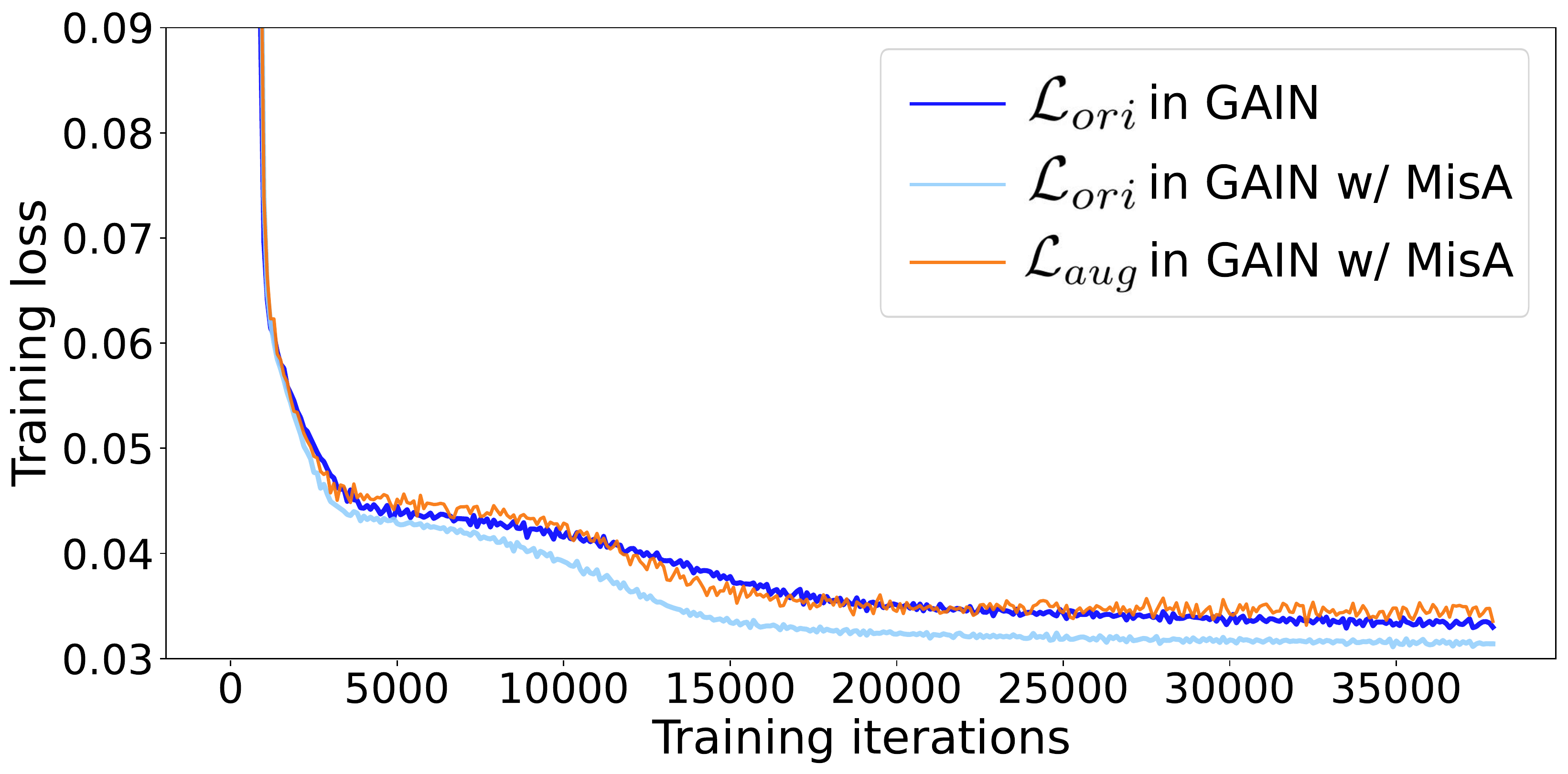}}}
\caption{Training loss curves of  the vanilla GAIN \cite{Yoon2018} and its augmented version.}
\label{LossCur}
\end{figure}

This fact is further illustrated by the loss curves at different trainining stages in Figure \ref{LossCur}.
We observed that, after a small training phase, the auxiliary loss based on the augmented samples begins to rapidly decrease,
driving the generator's loss to decrease faster and resulting in better imputation performance at the end of training.

\subsection{Sensitivity of the Regularization Coefficient}
In this subsection
we study the influence of the regularization coefficient $\alpha $ in \eqref{hybridloss}.
We test DAE \cite{Gondara2018}, GAIN \cite{Yoon2018},  VAEAC \cite{Nazabal2020} and MCFlow \cite{Richardson2020} on three UCI datasets,
the results are reported in Table \ref{UCI_RMSE_alpha}.

\begin{table}[!htb]
\center
\caption{ Comparison of RMSE between the baseline models and their corresponding augmented versions with various $\alpha $ on UCI datasets.}
\label{UCI_RMSE_alpha}
\scalebox{0.8}{
\begin{tabular}{cccccc}
\hline
            	              &$\alpha$            &Abalone            &Ionosphere         &Sonar            &Wine\\
\cline{2-6}
DAE \cite{Gondara2018}	         &-                   &0.1424             &0.2476             &0.2028           &0.2110\\
\hline
\multirow{4}*{DAE+}           &1                   &0.1344             &0.2300             &0.1932           &0.2006\\
                              &5		        &\textbf{0.1292}    &0.2410             &\textbf{0.1919}  &0.2015\\
                              &20                  &0.138              &0.2426             &0.2017           &\textbf{0.1988}\\
                              &50                  &0.1351             &\textbf{0.2176}    &0.2089           &0.2003\\
\hline
    GAIN \cite{Yoon2018}       &-                   &0.1597             &0.2941             &0.3534           &0.2476\\
\hline
\multirow{4}*{GAIN+}          &10                  &0.1214             &0.2658             &0.2820           &0.2148\\
                              &50                  &0.1133             &0.2531             &0.2542           &\textbf{0.1950}\\
				         &100	              &\textbf{0.1067}   &0.2509	           &\textbf{0.2480}  &0.2170\\
                              &200                 &0.1119             &\textbf{0.2410}    &0.2549           &0.2119\\
\hline
    MCFlow \cite{Richardson2020} &-                 &0.0880             &0.2311             &0.1613           &0.2057\\
\hline
\multirow{4}*{MCFlow+ }       &1                   &0.0871             &0.2272             &\textbf{0.1511}  &\textbf{0.1991}\\
                              &5                   &\textbf{0.0868}    &0.2265             &0.1518           &0.2050\\
				         &20	              &0.0874             &\textbf{0.2261}    &0.1533           &0.2107\\
                              &50                  &0.0879             &0.2295             &0.1546           &0.2133\\
\hline
    VAEAC \cite{Nazabal2020}   &-                   &0.1231             &0.2419             &0.2309           &0.1995\\
\hline
\multirow{4}*{VAEAC+}         &10	              &0.1176             &0.2311             &0.2172           &0.1936\\
                              &50	              &\textbf{0.0919}    &0.2282             &\textbf{0.1923}  &0.1899\\
                              &100                 &0.1148              &\textbf{0.2281}    &0.2041           &0.1949\\
                              &200                 &0.1145              &0.2317             &0.2097           &\textbf{0.1847}\\
\hline
\end{tabular}}
\end{table}

%\begin{table}[!htb]
%\center
%\caption{Comparison of PNSR between  MisGAN and its corresponding augmented version with various $\alpha $ on MNIST and CelebA datasets.}
%\label{CelebA_PSNR_alpha}
%\scalebox{0.8}{
%\begin{tabular}{cccc}
%	\hline
%                   	          &$\alpha$  	   &MNIST               &CelebA\\
%\cline{2-4}
%    MisGAN \cite{Li2019}        &-           &19.97	            &25.80\\
%\hline
%\multirow{4}*{MisGAN+}          &1             &\textbf{21.08}	    &26.61\\
%                               &10            &20.88	            &\textbf{27.26}\\
%                                &50            &20.34	            &26.58\\
%                                &100           &20.07	            &25.64\\
%\hline
%\end{tabular}}
%\end{table}

As seen from Table \ref{UCI_RMSE_alpha},
Missingness Augmentation approach  can consistently improve the original models under a wide rage of regularization coefficients.
An interesting phenomenon is that even when the value of $ \alpha $  is large,
which means that the training process mainly focuses on the reconstruction loss $\mathcal{L}_\mathrm{aug}$
and pays little attention to the original loss $ \mathcal{L} $,
our method still achieves a good results, which implicitly  indicates the importance of the reconstruction loss.

\subsection{Performance under MAR and MNAR Mechanisms}
To investigate the performance of our method under different missing mechanisms,
we  take the Ionosphere and Wine datasets as examples and  generate MAR and MNAR data as in \cite{Muzellec2020}.
Under the MAR setting, a fixed subset of variables that cannot have missing values is sampled for each experiment.
The remaining variables are then assigned missing values according to a logistic model with random weights,
which uses the non-missing variables as inputs.
A bias term is fitted using line search to achieve the desired proportion of missing values.
In the MNAR setting, the same approach as the MAR mechanism is used, but the inputs of the logistic model are masked by an MCAR mechanism.

\label{Performance on MAR and MNAR data}
\begin{table}[!htb]
\center
\caption{Comparison of RMSE between the baseline models and their corresponding augmented versions  under MAR and MNAR mechanisms. }
\label{UCI_MAR_MNAR}
\scalebox{0.9}{
\begin{tabular}{ccccccc}
	\hline
                                      &\multicolumn{3}{c}{MAR}                                    &\multicolumn{3}{c}{MNAR}  \\
\cline{2-7}
                                      &Abalone            &Ionosphere         &Wine               &Abalone             &Ionosphere          &Wine\\
\hline
    DAE \cite{Gondara2018}            &0.1768             &0.2944             &0.2744             &0.1991              &0.3286              &0.3117\\
    DAE+                              &\textbf{0.1516}    &\textbf{0.2718}    &\textbf{0.2467}    &\textbf{0.1620}     &\textbf{0.2857}     &\textbf{0.2683}\\
\hline
    DAEMA \cite{Tihon2021}            &\textbf{0.1349}    &\textbf{0.2891}    &0.2513             &\textbf{0.1467}     &0.3281              &0.2648  \\
    DAEMA+                            &0.1393             &0.2929             &\textbf{0.2461}    &0.1505              &\textbf{0.3272}     &\textbf{0.2566}\\
\hline
    GAIN \cite{Yoon2018}              &0.1647             &0.3083             &0.2515             &0.1749              &0.3167              &0.2759\\
    GAIN+                             &\textbf{0.1105}    &\textbf{0.2665}    &\textbf{0.2193}    &\textbf{0.1220}     &\textbf{0.2783}     &\textbf{0.2561}\\
\hline
    MCFlow \cite{Richardson2020}      &0.0955             &0.2447             &0.2291             &0.1015              &0.2649              &0.2392\\
    MCFlow+                           &\textbf{0.0951}    &\textbf{0.2379}    &\textbf{0.2272}    &\textbf{0.1010}     &\textbf{0.2589}     &\textbf{0.2353}\\
\hline
    VAEAC \cite{Nazabal2020}          &0.1436             &0.2725             &0.2413             &0.1519              &0.2983              &\textbf{0.2464}\\
    VAEAC+                            &\textbf{0.1080}    &\textbf{0.2574}    &\textbf{0.2332}    &\textbf{0.1201}     &\textbf{0.2951}     &0.2510\\
\hline
\end{tabular}}
\end{table}

Table \ref{UCI_MAR_MNAR} shows that our approach can still achieve consistent performance improvements  in MAR and MNAR scenarios,
with the exception of DAEMA and VAEAC on a few datasets.
This is probably because that DAEMA \cite{Tihon2021} and VAEAC \cite{Muzellec2020} themselves are very sensitive to the distribution of missingness,
but here the missingness distributions of the augmented samples and the original data are quite different.
Constructing augmented samples that conform to the actual missing mechanism would likely resolve this issue.

\section{Conclusion}
Missingness Augementation (MisA) is a simple but general data augmentation approach
 that leverages the model outputs to generate augmented incomplete samples on the fly, thereby enhancing the quality of imputation.
This approach is suitable not only for highly structured homogeneous data, such as images, but also for common heterogeneous datasets, such as the UCI \cite{Dua2017}.
We evaluated MisA on several benchmark generative imputation models, and all achieved significant improvements.
In addition, MisA exhibited superior performance under various data missing mechanisms, thus demonstrating its practicality.
Our study offers important insights into the development of precise and efficient methodologies for missing data imputation,
which hold promise for substantial benefits in diverse real-world scenarios.

\bigskip
\textbf{Acknowledgments}. This research is partially supported by National Natural Science Foundation of China (11771257) and Natural Science Foundation of Shandong Province (ZR2021MA010, ZR2022QF064).

\end{document}